\documentclass[11pt]{article}
\usepackage[utf8]{inputenc}
\usepackage[T1]{fontenc}
\usepackage{lmodern}
\usepackage{booktabs}
\usepackage{longtable}
\usepackage{array}
\usepackage{graphicx}
\usepackage{hyperref}
\usepackage{geometry}
\providecommand{\tightlist}{%
  \setlength{\itemsep}{0pt}\setlength{\parskip}{0pt}}

\title{Staged Factorial Screening for Budget-Constrained Micro-Pretraining}
\author{{\large Felipe Chavarro Polania}\\[0.35em]{\normalsize Hewlett Packard Enterprise}\\[0.15em]{\small\texttt{felipechavarropolania@hpe.com}}}
\date{2026-04-27}

\begin{document}
\maketitle

\begin{abstract}
Budget-constrained micro-pretraining is common in automated research
loops because many candidate recipes must be triaged on shared
accelerators before larger search budgets are spent. Best-so-far
trajectories can find better recipes, but they do not identify which
factors drive early performance differences. We test whether a staged
fractional-factorial workflow can recover stable early effect structure
under strict wall-clock budgets.

On a fixed autoresearch-derived single-GPU training loop we run
\texttt{613} experiments across pilot and follow-up screens at
\texttt{2}, \texttt{5}, and \texttt{10} minutes; full
\texttt{16}-condition seeded reruns at \texttt{5} and \texttt{10}
minutes; targeted anchor checks; same-host greedy and matched-cost
random baselines; a \texttt{60}-minute bridge package; and bounded
Windows A100 and Linux L40S anchor continuations through \texttt{24h}.
Main penalties from total batch, depth, and width are largest at short
budgets and relax as budget increases. Within the predeclared seeded
full-screen families, \texttt{D}, \texttt{A}, \texttt{B}, and \texttt{C}
retain non-zero estimates at \texttt{5} and \texttt{10} minutes after
within-budget Benjamini-Hochberg correction, while \texttt{E} does not.
Rerun-complete D-fixed follow-up analysis keeps interactions present but
smaller in absolute magnitude than the main penalties. Random search can
reach strong incumbents in this \texttt{32}-condition space, but
repeatedly in the same low-penalty region and without factor
attribution. The \texttt{60}-minute bridge anchor is best, although that
package does not separate workflow refinement from the larger bridge
model\textquotesingle s capacity advantage. In bounded \texttt{12h} and
\texttt{24h} three-anchor continuations on both hosts, the bridge has
the lowest sample mean while the non-bridge ordering stays
host-sensitive.

We therefore present a bounded methods result: use short designed
screens to identify high-penalty directions, confirm promising anchors
under repeated runs, and refine locally inside the reduced space. The
evidence supports a bridge-centered recommendation through \texttt{24h}
on two hosts, not hardware-invariant ranking or general
hyperparameter-optimization superiority.
\end{abstract}

\section{1. Introduction}\label{1-introduction}

Automated training workflows make it easy to launch many short runs, but
they do not make early-training behavior easier to interpret. A
best-so-far trajectory can locate a better recipe without identifying
whether the gain came from depth, width, batch size, learning rate, or a
specific interaction among them. If the goal is early structural
understanding rather than only incumbent improvement, experimental
design matters.

This paper studies that design problem in the budget-constrained
micro-pretraining regime. We ask whether a compact staged factorial
screen can identify stable early penalty directions under strict
wall-clock budgets, and whether the resulting signals remain strong
enough to support a practical screen-then-refine workflow. The focus is
not long-horizon convergence or full cross-hardware equivalence. The
focus is what can be learned early, cheaply, and reproducibly on a
runnable training loop before larger search budgets are committed, with
a bounded later cross-host check used to test whether the strongest
anchor-level signal remains visible outside the original runtime path.

We test two bounded hypotheses on the main host. \texttt{H1}: the
dominant early main-effect penalties from total batch and model size
relax materially as wall-clock budget increases from \texttt{2} to
\texttt{10} minutes. \texttt{H2}: after screening and local bridge
refinement, the reduced centered-width region remains informative at
\texttt{60} minutes and later anchor-level continuations by preserving
separation from the predeclared control on the main host and staying
competitive with or better than the bounded greedy incumbent.

Our evidence supports a narrow methodological conclusion. Under
\texttt{2}- to \texttt{10}-minute budgets on the main host, batch and
model-size penalties dominate the early effect structure and relax
substantially as budget increases. Full seeded reruns of the \texttt{5}-
and \texttt{10}-minute screens show that \texttt{D}, \texttt{A},
\texttt{B}, and \texttt{C} retain non-zero estimates after within-budget
BH-FDR correction while \texttt{E} does not. Rerun-complete follow-up
analysis shows that interactions remain real but secondary in absolute
size. A targeted multi-seed confirmation layer shows that budget and
condition structure dominate seed identity in the selected anchor
subset. A matched-cost random baseline shows that strong incumbents can
also be found by chance in this small space, but mostly by landing in
the same low-penalty region identified by the screen. A later
\texttt{60}-minute bridge package over four predeclared anchors shows
that the reduced-space bridge region remains operationally informative
at a longer horizon, subject to an unresolved capacity confound inside
that anchor set. Still later \texttt{12h} and \texttt{24h} three-anchor
continuations on Windows and Linux keep the bridge anchor at the lowest
sample mean on both hosts while again showing that the rest of the
anchor hierarchy remains host-sensitive. Taken together, these results
support a staged workflow for this regime: screen early, confirm anchor
regions, then refine locally inside the reduced space.

\section{2. Contributions}\label{2-contributions}

This paper makes three contributions.

\begin{enumerate}
\def\labelenumi{\arabic{enumi}.}
\tightlist
\item
  It presents a staged short-horizon screening methodology for
  micro-pretraining: factorial screen, focused confirmation, and local
  refinement inside a reduced space.
\item
  It provides host-bounded empirical evidence that early recipe effects
  are strongly budget-dependent, with main penalties from total batch
  and model size relaxing substantially between \texttt{2} and
  \texttt{10} minutes.
\item
  It demonstrates a bounded empirical pattern in this regime:
  \texttt{D}, \texttt{A}, \texttt{B}, and \texttt{C} retain seeded
  short-budget effects while \texttt{E} does not; random search reaches
  competitive incumbents without factor attribution; and the
  bridge-centered anchor has the lowest sample mean through small
  dual-host \texttt{12h} and \texttt{24h} continuations even though
  non-bridge ordering remains host-sensitive.
\end{enumerate}

\section{3. Related Work}\label{3-related-work}

Autoresearch-style systems focus on fast experiment throughput and
best-so-far progress rather than designed attribution {[}1, 2{]}.

Classical hyperparameter search establishes the baseline comparison set.
Random search is a strong default in high-dimensional hyperparameter
spaces {[}9{]}, and later work on functional ANOVA and random-forest
surrogates estimates which hyperparameters matter after search data have
already been collected {[}10{]}. Those papers motivate our emphasis on
attribution, but they do not study compact staged screening inside a
fixed short-horizon training loop.

Budgeted hyperparameter optimization methods such as Hyperband and BOHB
allocate resources adaptively across configurations to improve anytime
incumbent quality under finite budgets {[}11, 12{]}. Bayesian
optimization provides another major optimizer-centered line of
comparison {[}13{]}. Population-based training and asynchronous
successive halving are also natural adaptive alternatives for
distributed hyperparameter selection {[}20, 21{]}. Their target is
efficient optimizer performance over collections of tasks. Our target is
different: direct factor readout on one runnable host before a larger
automated search budget is committed.

Classical design-of-experiments provides the screening logic behind this
paper. Fractional-factorial designs trade aliasing against coverage in
order to expose large main effects and a controlled subset of
interactions early {[}14, 15, 16{]}. Response-surface methodology and
central-composite-style follow-ups are natural DOE extensions once a
promising region is found {[}17, 18{]}. We apply the screening logic to
micro-pretraining, then add a confirmation layer and a longer-horizon
bridge package tailored to the short-budget training workflow.

Large-scale scaling and optimizer studies provide important context but
ask different questions: cross-scale trends, compute-optimal training
rules, transfer rules, and architecture-aware defaults {[}3, 4, 19{]},
optimizer behavior under larger training horizons {[}5, 6, 7, 8{]}, or
learning-rate schedule diagnostics such as cyclical learning rates and
super-convergence {[}22, 23{]}. The gap we address is narrower. We ask
whether a staged designed screen can recover useful factor structure
early enough to reduce the search space before heavier automated search
begins on the same host-bounded training loop.

\section{4. Experimental Setup}\label{4-experimental-setup}

\subsection{4.1 Platform and Baseline}\label{41-platform-and-baseline}

The main study runs execute on a remote Windows host with one NVIDIA
A100 40GB GPU. The harness is an autoresearch seed baseline adapted to
this runtime path (SDPA fallback and disabled \texttt{torch.compile}).
We treat this as the primary measured environment, not an implementation
footnote. A later bounded replication package is run on a separate Linux
host with one NVIDIA L40S GPU in order to test whether the
bridge-centered anchor result remains visible after a host,
operating-system, and accelerator change. The Linux package is
intentionally not a like-for-like second-A100 replication; it is a
bounded portability check on the available independent Linux
accelerator, so positive transfer supports directional robustness rather
than matched-hardware equivalence.

The original screening, greedy, bridge, and D-fixed rerun blocks were
executed under an earlier fixed-seed harness. For the later confirmation
layers we added explicit \texttt{RUN\_SEED} support and ran both a
targeted \texttt{90}-run anchor subset across \texttt{2}, \texttt{5},
and \texttt{10} minutes and full \texttt{16}-condition seeded screens at
\texttt{5} and \texttt{10} minutes (\texttt{160} rows total). The paper
therefore contains both legacy fixed-seed blocks and later multi-seed
packages: seed variation is quantified directly on selected anchors
across all three budgets and on the full screening design at \texttt{5}
and \texttt{10} minutes.

The frozen runner uses locally cached training shards from
\texttt{karpathy/climbmix-400b-shuffle}, with
\texttt{shard\_06542.parquet} pinned as the validation shard.
Tokenization uses a rustbpe-trained, tiktoken-compatible BPE with
vocabulary \texttt{8192}. All runs use context length \texttt{2048} and
report final \texttt{val\_bpb} over \texttt{40\ *\ 524288} validation
tokens from the pinned shard. Appendix A records the dataset URL, shard
identifier, tokenizer artifacts, source snapshot, and reproducibility
bundle contents.

Model families are parameterized by \texttt{DEPTH} and
\texttt{ASPECT\_RATIO} on a Transformer-style causal language-model
runner {[}24{]}. For a given depth, the nominal width is
\texttt{DEPTH\ *\ ASPECT\_RATIO}, rounded up to the next multiple of
\texttt{HEAD\_DIM=128}; the attention head count is then
\texttt{n\_embd\ /\ 128}, and the attention window pattern is fixed to
\texttt{SSSL} with the last layer forced to full context. Optimization
is also frozen except for the experimental factors: matrix-valued
transformer weights use Muon, embeddings, unembedding, and scalar
parameters use AdamW, AdamW betas are \texttt{(0.8,\ 0.95)}, warmup
ratio is fixed to \texttt{0.0}, and the Windows fallback path uses
device batch size \texttt{32} with gradient accumulation to realize the
configured \texttt{TOTAL\_BATCH\_SIZE}.

Primary outcome is final validation bits per byte (\texttt{val\_bpb},
lower is better), computed as per-token cross-entropy in nats on the
pinned validation stream divided by UTF-8 target bytes and converted to
bits per byte; special tokens are excluded from both sums.

\subsection{4.2 Factors}\label{42-factors}

Five pilot factors:

\begin{longtable}[]{@{}llll@{}}
\toprule
Code & Factor & Low & High \\
\midrule
\endhead
\bottomrule
\endlastfoot
\texttt{A} & \texttt{DEPTH} & 6 & 8 \\
\texttt{B} & \texttt{ASPECT\_RATIO} & 48 & 72 \\
\texttt{C} & \texttt{MATRIX\_LR} & 0.03 & 0.05 \\
\texttt{D} & \texttt{TOTAL\_BATCH\_SIZE} & 262144 & 524288 \\
\texttt{E} & \texttt{WARMDOWN\_RATIO} & 0.25 & 0.50 \\
\end{longtable}

\subsection{4.3 Screening Design
Definition}\label{43-screening-design-definition}

The pilot screen is a regular \texttt{2\^{}(5-1)} fractional-factorial
design with generator \texttt{E\ =\ A*B*C*D}, equivalently defining
relation \texttt{I\ =\ A*B*C*D*E}. This is a resolution-V design. Main
effects are therefore aliased only with four-factor terms, while each
two-factor term is aliased with a complementary three-factor term. For
example, \texttt{A} is aliased with \texttt{BCDE}, \texttt{B} with
\texttt{ACDE}, and \texttt{A:B} with \texttt{CDE}.

We use the pilot screen strictly as a main-effect screening design. The
paper does not claim that the initial \texttt{16}-run pilot by itself
identifies unaliased two-factor interactions. Interaction discussion is
moved to the separate D-fixed follow-up package, where the design and
model are tailored to the reduced factor set.

\subsection{4.4 Statistical Methods}\label{44-statistical-methods}

All modeled factors use coded levels \texttt{\{-1,\ +1\}}. Under this
coding, a high-minus-low effect is \texttt{2*beta}, where \texttt{beta}
is the fitted coefficient in the corresponding linear model.

Condition-level means and \texttt{95\%} confidence intervals are
reported from repeated runs within each cell using Student-\texttt{t}
intervals based on the sample standard deviation and the within-cell
replicate count. These intervals appear in the seeded condition
summaries, the \texttt{60}-minute bridge package, the Linux cross-host
anchor package, and the later \texttt{12h} and \texttt{24h} anchor
continuations.

For the seeded full-screen reruns, we fit separate budget-specific
ordinary least-squares models

\texttt{val\_bpb\ \textasciitilde{}\ A\ +\ B\ +\ C\ +\ D\ +\ E\ +\ seed\_factor}

at \texttt{5} and \texttt{10} minutes. We report high-minus-low effects,
two-sided coefficient \texttt{p}-values, and \texttt{95\%} Wald
intervals using the residual degrees of freedom from each fitted model.
For these seeded full-screen main effects, we additionally apply
Benjamini-Hochberg false-discovery-rate control within each predeclared
five-effect family at a fixed budget. At both \texttt{5} and \texttt{10}
minutes, \texttt{D}, \texttt{A}, \texttt{B}, and \texttt{C} survive
BH-FDR at \texttt{q=0.05}, while \texttt{E} does not. In this paper,
"retain a non-zero estimate" for the seeded full screens means that the
reported \texttt{95\%} interval excludes zero and the corresponding main
effect also survives that within-budget BH correction.

As a sensitivity check on the correction family, applying BH-FDR once
across the combined \texttt{10} seeded full-screen main-effect tests
(\texttt{5} factors x \texttt{2} budgets) gives the same qualitative
retention set: \texttt{D}, \texttt{A}, \texttt{B}, and \texttt{C}
survive at both budgets, while \texttt{E} does not. We retain the
within-budget presentation because the models and decision questions are
budget-specific, but the conclusion does not depend on that split.

For the targeted seed-confirmation subset, we fit the fixed-effects
model

\texttt{val\_bpb\ \textasciitilde{}\ C(budget\_factor)\ *\ C(condition\_factor)\ +\ C(seed\_factor)}

and summarize variance shares with Type-II ANOVA
\texttt{eta\_sq\ =\ SS\_term\ /\ SS\_total}. These \texttt{eta\_sq}
values are descriptive effect-size summaries, not variance-component
estimates from a random-effects model.

For the D-fixed follow-up, we analyze the legacy fixed-seed regime
separately from the later explicit-seed packages. The pooled follow-up
model is

\texttt{val\_bpb\ \textasciitilde{}\ budget10\ +\ A\ +\ B\ +\ C\ +\ E\ +\ A:B\ +\ A:C\ +\ B:C\ +\ A:E\ +\ B:E\ +\ C:E\ +\ budget10:(A\ +\ B\ +\ C\ +\ E\ +\ A:B\ +\ A:C\ +\ B:C\ +\ A:E\ +\ B:E\ +\ C:E)}

using only the original and rerun D-fixed blocks from that same regime.
The reported intervals are model-based Wald intervals from the fitted
covariance matrix and should be read as rerun-variability summaries
within the fixed-seed regime, not as independent-seed inference.

Because this block lacks independent seed variation, its intervals are
expected to understate the variability a broad independently seeded
interaction study would see. We therefore use the D-fixed block only to
diagnose interaction structure inside the legacy regime, not as a
multiplicity-corrected decision layer.

The pairwise dominance counts (\texttt{100/100}, \texttt{16/16}, and
similar) are descriptive cross-seed win counts over all left-right seed
products, with denominator \texttt{n\_left\ *\ n\_right}. Because each
seed value participates in multiple pairings, we do not treat those
denominators as independent Bernoulli trials, and we do not attach
\texttt{p}-values or confidence intervals to them.

We report raw two-sided \texttt{p}-values for transparency. Outside the
seeded full-screen main-effect families, these \texttt{p}-values are
descriptive diagnostics rather than multiplicity-corrected decision
rules.

\subsection{4.5 Run Accounting and
Reproducibility}\label{45-run-accounting-and-reproducibility}

Table 1 enumerates the completed packages that feed the present
manuscript. Using this accounting, the paper currently draws on
\texttt{613} completed runs: \texttt{569} on the Windows A100 host and
\texttt{44} on the Linux L40S host.

\begin{longtable}[]{@{}llllr@{}}
\toprule
package & host & regime & budgets & rows \\
\midrule
\endhead
\bottomrule
\endlastfoot
pilot screening & Windows A100 & legacy fixed-seed & \texttt{2},
\texttt{5}, \texttt{10} min & \texttt{48} \\
extreme-condition replication & Windows A100 & legacy fixed-seed &
\texttt{2}, \texttt{10} min & \texttt{20} \\
D-fixed follow-up original + rerun & Windows A100 & legacy fixed-seed &
\texttt{5}, \texttt{10} min & \texttt{64} \\
same-host greedy baselines & Windows A100 & legacy fixed-seed &
\texttt{5}, \texttt{10} min & \texttt{70} \\
centered-width bridge probes & Windows A100 & legacy fixed-seed &
\texttt{2}, \texttt{5}, \texttt{10} min & \texttt{9} \\
targeted seed confirmation & Windows A100 & explicit seeded reruns &
\texttt{2}, \texttt{5}, \texttt{10} min & \texttt{90} \\
full seeded screens & Windows A100 & explicit seeded reruns &
\texttt{5}, \texttt{10} min & \texttt{160} \\
random-search baseline & Windows A100 & matched-cost independent draws &
\texttt{10} min & \texttt{80} \\
60-minute bridge package & Windows A100 & explicit seeded reruns &
\texttt{60} min & \texttt{16} \\
Linux cross-host anchors & Linux L40S & explicit seeded reruns &
\texttt{10}, \texttt{60} min & \texttt{32} \\
Windows 12h anchors & Windows A100 & explicit seeded reruns &
\texttt{12} h & \texttt{6} \\
Linux 12h anchors & Linux L40S & explicit seeded reruns & \texttt{12} h
& \texttt{6} \\
Windows 24h anchors & Windows A100 & explicit seeded reruns &
\texttt{24} h & \texttt{6} \\
Linux 24h anchors & Linux L40S & explicit seeded reruns & \texttt{24} h
& \texttt{6} \\
\textbf{total} & & & & \textbf{\texttt{613}} \\
\end{longtable}

Every package is defined by an explicit matrix or result table with
factor codings, decoded hyperparameters, time budget, and seed
identifiers where applicable. The manuscript therefore distinguishes
data-generation regimes at the package level instead of treating all
completed runs as one exchangeable sample. We freeze the upstream
baseline commit and the exact local source snapshot used for this branch
in Appendix A, together with the package-level matrix files and
machine-readable summaries that define the experimental configurations.
The later dual-host \texttt{24h} packages are treated as bounded
three-anchor hardening results rather than as broad reruns of the full
design space.

\subsection{4.6 Experimental Sequence}\label{46-experimental-sequence}

The study follows ten stages:

\begin{enumerate}
\def\labelenumi{\arabic{enumi}.}
\tightlist
\item
  \textbf{Pilot screening} across \texttt{2}, \texttt{5}, and
  \texttt{10} minute budgets to estimate the dominant early penalties.
\item
  \textbf{Reduced follow-up reruns} to test whether interaction
  structure remains after fixing a major penalty source.
\item
  \textbf{Bounded greedy comparison} to contrast best-so-far search with
  the structural readout from designed screening.
\item
  \textbf{Matched-cost random-search baseline} over five independent
  \texttt{16}-draw batches to test whether competitive incumbents appear
  routinely without structured attribution.
\item
  \textbf{Targeted seed confirmation} over extremes and bridge points to
  test whether the main ordering holds across independent seeds in a
  compact anchor subset.
\item
  \textbf{Full seeded-screen reruns} over all \texttt{16} screening
  conditions at \texttt{5} and \texttt{10} minutes to attach seed
  uncertainty to the main-effect estimates and condition ordering.
\item
  \textbf{Longer-horizon bridge package} over four predeclared anchors
  at \texttt{60} minutes to test whether the reduced-space ranking
  remains informative beyond the short screening horizon.
\item
  \textbf{Cross-host Linux anchor replication} over the same four
  anchors at \texttt{10} and \texttt{60} minutes to test whether the
  bridge-centered signal remains visible outside the original Windows
  A100 path.
\item
  \textbf{Dual-host \texttt{12h} anchor continuations} over
  \texttt{bridge}, \texttt{greedy}, and \texttt{control} to test whether
  the bridge keeps the lowest sample mean in these small descriptive
  three-anchor continuations on both hosts.
\item
  \textbf{Dual-host \texttt{24h} anchor continuations} over the same
  three anchors to test whether that bridge-centered sample-mean result
  persists at a full-day horizon on both hosts.
\end{enumerate}

\section{5. Results}\label{5-results}

\subsection{5.1 Budget Curves from Pilot and Seeded Full
Screens}\label{51-budget-curves-from-pilot-and-seeded-full-screens}

Main-effect estimates (high minus low) on \texttt{val\_bpb}. The
\texttt{2}-minute column is the original single-seed pilot estimate and
is used descriptively. The \texttt{5}- and \texttt{10}-minute columns
are seeded fixed-effects estimates from the full \texttt{16}-condition
reruns (\texttt{5} seeds per condition).

Pilot \texttt{2}-minute estimates:

\begin{longtable}[]{@{}lr@{}}
\toprule
factor & effect \\
\midrule
\endhead
\bottomrule
\endlastfoot
\texttt{A} & \texttt{+0.1112} \\
\texttt{B} & \texttt{+0.1109} \\
\texttt{C} & \texttt{-0.0597} \\
\texttt{D} & \texttt{+0.1847} \\
\texttt{E} & \texttt{+0.0448} \\
\end{longtable}

Seeded \texttt{5}-minute estimates:

\begin{longtable}[]{@{}lrlrr@{}}
\toprule
factor & effect & 95\% CI & p & BH q \\
\midrule
\endhead
\bottomrule
\endlastfoot
\texttt{A} & \texttt{+0.0568} & {[}\texttt{+0.0426}, \texttt{+0.0710}{]}
& \texttt{1.98e-11} & \texttt{4.94e-11} \\
\texttt{B} & \texttt{+0.0441} & {[}\texttt{+0.0299}, \texttt{+0.0583}{]}
& \texttt{3.57e-08} & \texttt{5.94e-08} \\
\texttt{C} & \texttt{-0.0165} & {[}\texttt{-0.0307}, \texttt{-0.0023}{]}
& \texttt{0.0232} & \texttt{0.0290} \\
\texttt{D} & \texttt{+0.0818} & {[}\texttt{+0.0676}, \texttt{+0.0960}{]}
& \texttt{9.01e-18} & \texttt{4.50e-17} \\
\texttt{E} & \texttt{+0.0054} & {[}\texttt{-0.0088}, \texttt{+0.0196}{]}
& \texttt{0.454} & \texttt{0.454} \\
\end{longtable}

Seeded \texttt{10}-minute estimates:

\begin{longtable}[]{@{}lrlrr@{}}
\toprule
factor & effect & 95\% CI & p & BH q \\
\midrule
\endhead
\bottomrule
\endlastfoot
\texttt{A} & \texttt{+0.0180} & {[}\texttt{+0.0141}, \texttt{+0.0218}{]}
& \texttt{8.86e-14} & \texttt{2.21e-13} \\
\texttt{B} & \texttt{+0.0059} & {[}\texttt{+0.0021}, \texttt{+0.0098}{]}
& \texttt{0.00319} & \texttt{0.00532} \\
\texttt{C} & \texttt{-0.0045} & {[}\texttt{-0.0084}, \texttt{-0.0007}{]}
& \texttt{0.0219} & \texttt{0.0274} \\
\texttt{D} & \texttt{+0.0346} & {[}\texttt{+0.0308}, \texttt{+0.0385}{]}
& \texttt{9.03e-28} & \texttt{4.52e-27} \\
\texttt{E} & \texttt{-0.0013} & {[}\texttt{-0.0052}, \texttt{+0.0026}{]}
& \texttt{0.501} & \texttt{0.501} \\
\end{longtable}

Interpretation:

\begin{itemize}
\tightlist
\item
  \texttt{D} remains the largest penalty at \texttt{5} and \texttt{10}
  minutes, with a strong but incomplete relaxation as budget increases.
\item
  \texttt{A} remains the second-largest penalty at both seeded budgets
  and is clearly non-zero.
\item
  \texttt{B} retains a non-zero estimate after within-budget correction
  at both seeded budgets, but by \texttt{10} minutes it is small
  relative to \texttt{A} and \texttt{D}.
\item
  \texttt{C} remains beneficial at both seeded budgets, again with clear
  relaxation.
\item
  \texttt{E} is not distinguishable from zero at either seeded budget,
  so the earlier single-seed signal does not survive broad reruns.
\end{itemize}

In practical metric terms, the seeded \texttt{10}-minute batch-size
penalty (\texttt{D}, \texttt{+0.0346} \texttt{val\_bpb}) is about six
times the width penalty (\texttt{B}, \texttt{+0.0059}) and about eight
times the learning-rate benefit magnitude
(\texttt{\textbar{}C\textbar{}=0.0045}). These comparisons are
within-metric effect-size context only: they help rank recipe choices on
validation compression, but they do not establish downstream task
impact.

Figure 1 collects those main-effect estimates across budgets and makes
the relaxation pattern visible at a glance: the largest short-budget
penalties shrink materially by \texttt{5} and \texttt{10} minutes, while
\texttt{E} collapses toward zero under seeded reruns.

\begin{figure}
\centering
\includegraphics[width=\linewidth,height=0.82\textheight,keepaspectratio]{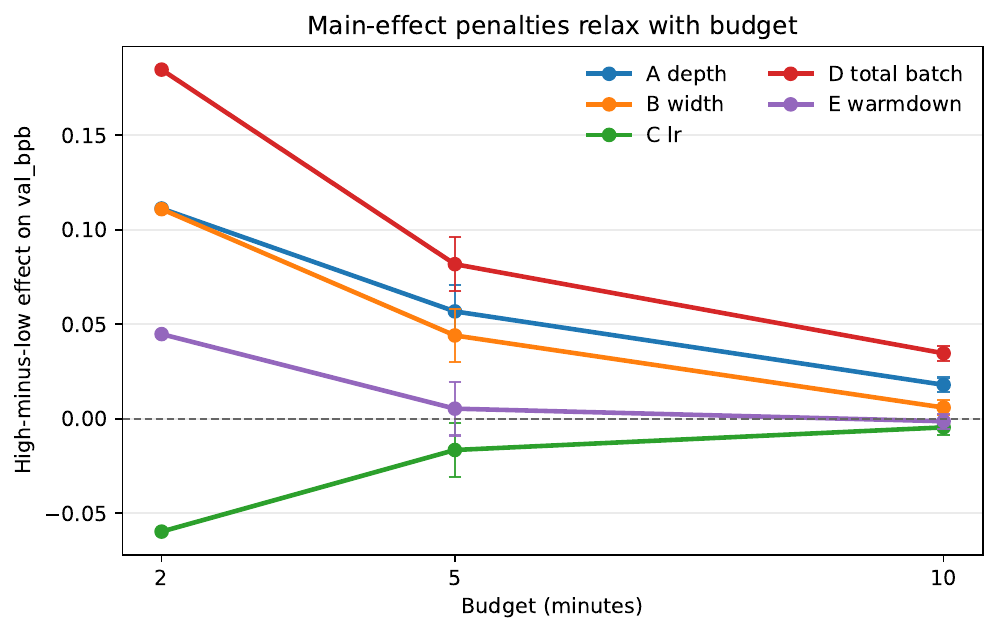}
\caption{Main-effect penalties relax with budget. The \texttt{2}-minute
points come from the original pilot; the \texttt{5}- and
\texttt{10}-minute points include seeded full-screen error bars.}
\end{figure}

\subsection{5.2 Replication at 2 and 10
Minutes}\label{52-replication-at-2-and-10-minutes}

2-minute extreme-condition replication (top 2 + bottom 2 from 2-minute
pilot):

\begin{longtable}[]{@{}lrrr@{}}
\toprule
source condition & n & mean \texttt{val\_bpb} & sd \\
\midrule
\endhead
\bottomrule
\endlastfoot
\texttt{1} & \texttt{2} & \texttt{1.317674} & \texttt{0.000163} \\
\texttt{3} & \texttt{2} & \texttt{1.274105} & \texttt{0.000267} \\
\texttt{14} & \texttt{2} & \texttt{1.719449} & \texttt{0.000733} \\
\texttt{16} & \texttt{2} & \texttt{1.675783} & \texttt{0.000358} \\
\end{longtable}

10-minute extreme-condition replication (top 4 + bottom 2 from 10-minute
pilot):

\begin{longtable}[]{@{}lrrr@{}}
\toprule
source condition & n & mean \texttt{val\_bpb} & sd \\
\midrule
\endhead
\bottomrule
\endlastfoot
\texttt{1} & \texttt{2} & \texttt{1.093624} & \texttt{0.000218} \\
\texttt{3} & \texttt{2} & \texttt{1.095242} & \texttt{0.000131} \\
\texttt{5} & \texttt{2} & \texttt{1.092018} & \texttt{0.000128} \\
\texttt{7} & \texttt{2} & \texttt{1.090640} & \texttt{0.000668} \\
\texttt{14} & \texttt{2} & \texttt{1.152082} & \texttt{0.000310} \\
\texttt{16} & \texttt{2} & \texttt{1.142647} & \texttt{0.000116} \\
\end{longtable}

Interpretation:

\begin{itemize}
\tightlist
\item
  Replications preserve strong separation between best and worst
  regions.
\item
  Variance is low, and top-vs-bottom separation is consistent within
  this sampled condition set.
\end{itemize}

\subsection{5.3 D-Fixed Follow-Up Reruns at 5 vs 10
Minutes}\label{53-d-fixed-follow-up-reruns-at-5-vs-10-minutes}

We analyze the D-fixed follow-up separately because it belongs to the
legacy fixed-seed regime. Within that regime, we combine the original
and full-rerun D-fixed ABCE blocks at \texttt{5} and \texttt{10} minutes
(\texttt{64} rows total; \texttt{32} per budget) in a pooled within-host
model over \texttt{A}, \texttt{B}, \texttt{C}, \texttt{E}, the six
two-way terms, and budget interactions. The intervals below quantify
rerun variability within that fixed-seed regime rather than
independent-seed uncertainty.

\begin{longtable}[]{@{}lll@{}}
\toprule
effect & 5 min effect {[}95\% CI{]} & 10 min effect {[}95\% CI{]} \\
\midrule
\endhead
\bottomrule
\endlastfoot
\texttt{A} & \texttt{+0.0227} {[}\texttt{+0.0214}, \texttt{+0.0240}{]} &
\texttt{+0.0084} {[}\texttt{+0.0071}, \texttt{+0.0097}{]} \\
\texttt{B} & \texttt{+0.0210} {[}\texttt{+0.0197}, \texttt{+0.0223}{]} &
\texttt{+0.0023} {[}\texttt{+0.0010}, \texttt{+0.0036}{]} \\
\texttt{C} & \texttt{-0.0085} {[}\texttt{-0.0098}, \texttt{-0.0072}{]} &
\texttt{-0.0013} {[}\texttt{-0.0026}, \texttt{0.0000}{]} \\
\texttt{E} & \texttt{+0.0014} {[}\texttt{+0.0001}, \texttt{+0.0027}{]} &
\texttt{-0.0028} {[}\texttt{-0.0041}, \texttt{-0.0015}{]} \\
\texttt{A:B} & \texttt{+0.0101} {[}\texttt{+0.0088}, \texttt{+0.0114}{]}
& \texttt{+0.0049} {[}\texttt{+0.0036}, \texttt{+0.0062}{]} \\
\texttt{A:C} & \texttt{-0.0054} {[}\texttt{-0.0067}, \texttt{-0.0041}{]}
& \texttt{-0.0012} {[}\texttt{-0.0024}, \texttt{+0.0001}{]} \\
\texttt{B:C} & \texttt{-0.0024} {[}\texttt{-0.0037}, \texttt{-0.0011}{]}
& \texttt{-0.0003} {[}\texttt{-0.0016}, \texttt{+0.0010}{]} \\
\texttt{A:E} & \texttt{+0.0017} {[}\texttt{+0.0004}, \texttt{+0.0030}{]}
& \texttt{-0.0005} {[}\texttt{-0.0018}, \texttt{+0.0008}{]} \\
\texttt{B:E} & \texttt{+0.0023} {[}\texttt{+0.0010}, \texttt{+0.0036}{]}
& \texttt{-0.0005} {[}\texttt{-0.0018}, \texttt{+0.0008}{]} \\
\texttt{C:E} & \texttt{-0.0018} {[}\texttt{-0.0031}, \texttt{-0.0005}{]}
& \texttt{-0.0009} {[}\texttt{-0.0022}, \texttt{+0.0004}{]} \\
\end{longtable}

Interpretation:

\begin{itemize}
\tightlist
\item
  Main effects shrink notably with budget, especially \texttt{A} and
  \texttt{B}.
\item
  \texttt{A:B} remains positive at both budgets, but its absolute effect
  declines from \texttt{+0.0101} to \texttt{+0.0049}; the pooled budget
  interaction for \texttt{A:B} is negative (\texttt{delta=-0.0052},
  \texttt{p=1.97e-06}).
\item
  Relative salience still rises because the main effects shrink faster:
  \texttt{\textbar{}A:B\textbar{}\ /\ \textbar{}A\textbar{}} increases
  from \texttt{0.444} at \texttt{5\ min} to \texttt{0.585} at
  \texttt{10\ min}.
\item
  Interactions remain secondary in absolute size even after the
  rerun-complete analysis.
\end{itemize}

\subsection{5.4 Same-Host Greedy
Baselines}\label{54-same-host-greedy-baselines}

Greedy summaries:

\begin{itemize}
\tightlist
\item
  5-minute greedy (\texttt{50} runs): \texttt{6} accepted updates, best
  \texttt{1.152694} at step \texttt{12}.
\item
  10-minute greedy (\texttt{20} runs): \texttt{9} accepted updates, best
  \texttt{1.090276} at step \texttt{15}.
\item
  10-minute greedy first hits \texttt{A=-1,\ D=-1} at step \texttt{7}.
\end{itemize}

Interpretation:

\begin{itemize}
\tightlist
\item
  Greedy improves incumbents effectively, especially early.
\item
  These trajectories show search behavior, but by themselves they do not
  identify which factors produced the gain; that attribution still comes
  from the designed comparisons.
\end{itemize}

\subsection{5.5 Matched-Cost Random Search
Baseline}\label{55-matched-cost-random-search-baseline}

We ran a matched-cost random baseline over the full
\texttt{32}-condition recipe space: five independent batches of
\texttt{16} random draws each at \texttt{10} minutes (\texttt{80} rows
total). This uses the same per-run budget as the original
\texttt{16}-run designed screen, but without structured coverage or
factor balancing.

Benchmarks from the original \texttt{10}-minute single-seed comparisons:

\begin{itemize}
\tightlist
\item
  pilot screened best: \texttt{1.090745}
\item
  greedy best: \texttt{1.090276}
\end{itemize}

Batch-best summary from the five random batches:

\begin{longtable}[]{@{}lr@{}}
\toprule
statistic & value \\
\midrule
\endhead
\bottomrule
\endlastfoot
mean batch best & \texttt{1.090615} \\
median batch best & \texttt{1.090450} \\
min batch best & \texttt{1.089706} \\
max batch best & \texttt{1.092171} \\
batches at or better than pilot screened best & \texttt{3\ /\ 5} \\
batches at or better than greedy best & \texttt{2\ /\ 5} \\
\end{longtable}

Per-batch winners:

\begin{longtable}[]{@{}lllr@{}}
\toprule
batch & best condition & recipe & best \texttt{val\_bpb} \\
\midrule
\endhead
\bottomrule
\endlastfoot
\texttt{1} & \texttt{full32\_c10} & \texttt{A-1\_B+1\_C-1\_D-1\_E+1} &
\texttt{1.090450} \\
\texttt{2} & \texttt{full32\_c06} & \texttt{A-1\_B-1\_C+1\_D-1\_E+1} &
\texttt{1.092171} \\
\texttt{3} & \texttt{full32\_c14} & \texttt{A-1\_B+1\_C+1\_D-1\_E+1} &
\texttt{1.089975} \\
\texttt{4} & \texttt{full32\_c14} & \texttt{A-1\_B+1\_C+1\_D-1\_E+1} &
\texttt{1.090771} \\
\texttt{5} & \texttt{full32\_c10} & \texttt{A-1\_B+1\_C-1\_D-1\_E+1} &
\texttt{1.089706} \\
\end{longtable}

Interpretation:

\begin{itemize}
\tightlist
\item
  In this small \texttt{32}-condition space, matched-cost random search
  can reach incumbents that are competitive with and sometimes slightly
  better than the original screened-best and greedy-best single-seed
  references.
\item
  This comparison is intentionally narrow: each random batch has
  \texttt{16} draws at the \texttt{10}-minute budget and is matched to a
  single-budget \texttt{16}-condition screen, not to the full
  multi-stage \texttt{613}-run evidence program.
\item
  The random baseline should therefore be read as an incumbent-quality
  stress test for the original \texttt{10}-minute screen, not as a
  cost-matched replacement for the full screen-confirm-refine workflow.
\item
  That result rules out a simple superiority claim based only on
  incumbent quality at \texttt{10} minutes. It does not test seeded
  mean-vs-mean superiority against the later rerun packages.
\item
  The random winners are not dispersed across the space. All five batch
  winners share \texttt{A=-1} and \texttt{D=-1}, and four of five also
  use \texttt{B=+1}. Random search succeeds mainly when it stumbles into
  the same low-penalty region that the designed screen isolates
  structurally.
\item
  The methodological value of the staged screen therefore remains
  attribution and disciplined refinement, not a guarantee that no random
  batch can hit a strong incumbent.
\end{itemize}

\subsection{5.6 Center-Point Bridge
Runs}\label{56-center-point-bridge-runs}

\texttt{ASPECT\_RATIO=64} bridge runs:

\begin{longtable}[]{@{}lrrr@{}}
\toprule
time budget & n & mean \texttt{val\_bpb} & sd \\
\midrule
\endhead
\bottomrule
\endlastfoot
\texttt{120} & \texttt{3} & \texttt{1.284982} & \texttt{0.000465} \\
\texttt{300} & \texttt{3} & \texttt{1.152658} & \texttt{0.000230} \\
\texttt{600} & \texttt{3} & \texttt{1.092224} & \texttt{0.000168} \\
\end{longtable}

Interpretation:

\begin{itemize}
\tightlist
\item
  Center-point mean performance improves as budget increases from 120s
  to 600s.
\item
  These runs improve comparability between greedy trajectories (which
  explored \texttt{64}) and two-level pilot screens
  (\texttt{48}/\texttt{72}).
\end{itemize}

\subsection{5.7 Targeted Seed Confirmation Across
Budgets}\label{57-targeted-seed-confirmation-across-budgets}

We ran a targeted seed-confirmation layer over six anchor conditions:
two prior best conditions, two prior worst conditions, and two bridge
conditions at \texttt{ASPECT\_RATIO=64}. This yields \texttt{90} rows
total (\texttt{6} conditions x \texttt{3} budgets x \texttt{5} seeds).
This targeted subset complements the later full seeded-screen reruns by
covering the \texttt{2}-minute regime and the centered-width bridge
conditions directly.

In a pooled fixed-effects model over budget, condition, and seed, the
dominant variance sources are budget and condition structure, not seed
identity:

\begin{longtable}[]{@{}lr@{}}
\toprule
term & \texttt{eta\_sq} \\
\midrule
\endhead
\bottomrule
\endlastfoot
budget & \texttt{0.6035} \\
condition & \texttt{0.2723} \\
budget:condition & \texttt{0.1079} \\
seed & \texttt{0.0044} \\
residual & \texttt{0.0118} \\
\end{longtable}

Descriptive cross-seed win counts over the seeded subset
(\texttt{n\_left\ *\ n\_right} pairings, not independent-trial
denominators):

\begin{longtable}[]{@{}lrrr@{}}
\toprule
budget & \texttt{best\textless{}worst} & \texttt{bridge\textless{}worst}
& \texttt{best\textless{}bridge} \\
\midrule
\endhead
\bottomrule
\endlastfoot
\texttt{2\ min} & \texttt{100/100} & \texttt{100/100} &
\texttt{78/100} \\
\texttt{5\ min} & \texttt{100/100} & \texttt{100/100} &
\texttt{62/100} \\
\texttt{10\ min} & \texttt{100/100} & \texttt{100/100} &
\texttt{25/100} \\
\end{longtable}

Local refinement signals from the same seeded subset:

\begin{longtable}[]{@{}lrr@{}}
\toprule
budget & \texttt{bridge\_d8\ -\ bridge\_d6} &
\texttt{bridge\_d6\ -\ best\_best} \\
\midrule
\endhead
\bottomrule
\endlastfoot
\texttt{2\ min} & \texttt{+0.1167} & \texttt{+0.0225} \\
\texttt{5\ min} & \texttt{+0.0191} & \texttt{-0.0019} \\
\texttt{10\ min} & \texttt{-0.0005} & \texttt{-0.0013} \\
\end{longtable}

Interpretation:

\begin{itemize}
\tightlist
\item
  Worst regions remain clearly separated from both best and bridge
  regions under explicit seed variation at all three budgets.
\item
  Seed effects are statistically non-zero in the targeted subset, but
  they are small relative to budget and condition structure.
\item
  The \texttt{ASPECT\_RATIO=64} bridges are operationally important: the
  depth-6 bridge overtakes the screened best extremes on mean
  \texttt{val\_bpb} at \texttt{5\ min}, and both bridges do so by
  \texttt{10\ min}.
\item
  At centered width, the depth penalty is large at \texttt{2\ min},
  smaller at \texttt{5\ min}, and near-tied by \texttt{10\ min}.
\end{itemize}

Figure 2 visualizes the same point from a variance-allocation
perspective: the seeded subset is driven primarily by budget and
condition structure, not by seed identity.

\begin{figure}
\centering
\includegraphics[width=\linewidth,height=0.82\textheight,keepaspectratio]{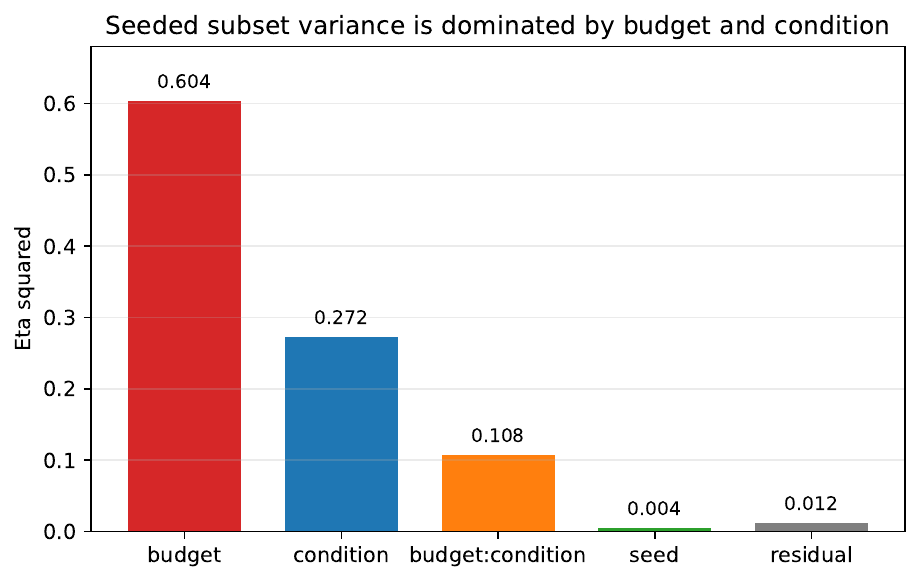}
\caption{In the seeded confirmation subset, variance is dominated by
budget and condition structure rather than seed identity.}
\end{figure}

\subsection{5.8 Full Seeded-Screen Confirmation at 5 and 10
Minutes}\label{58-full-seeded-screen-confirmation-at-5-and-10-minutes}

We then reran the full \texttt{16}-condition screening design at
\texttt{5} and \texttt{10} minutes with five independent seeds per
condition (\texttt{160} rows total). This is the main statistical
upgrade relative to the earlier draft because it removes the earlier
weakness that full-screen main effects at those budgets were single-seed
estimates.

Condition ordering in the seeded full screens:

For \texttt{5\ min}:

\begin{itemize}
\tightlist
\item
  best mean: \texttt{c07} = \texttt{1.160271} {[}\texttt{1.152515},
  \texttt{1.168028}{]}
\item
  second mean: \texttt{c03} = \texttt{1.163509} {[}\texttt{1.160124},
  \texttt{1.166894}{]}
\item
  worst mean: \texttt{c14} = \texttt{1.371836} {[}\texttt{1.307309},
  \texttt{1.436364}{]}
\end{itemize}

For \texttt{10\ min}:

\begin{itemize}
\tightlist
\item
  best mean: \texttt{c07} = \texttt{1.088859} {[}\texttt{1.085720},
  \texttt{1.091999}{]}
\item
  second mean: \texttt{c05} = \texttt{1.094040} {[}\texttt{1.090097},
  \texttt{1.097983}{]}
\item
  worst mean: \texttt{c14} = \texttt{1.164241} {[}\texttt{1.147242},
  \texttt{1.181241}{]}
\end{itemize}

Interpretation:

\begin{itemize}
\tightlist
\item
  The best region from the original screens holds under the seeded
  reruns: \texttt{c07} remains the best mean condition at both
  \texttt{5} and \texttt{10} minutes.
\item
  The worst region also holds: \texttt{c14} and \texttt{c16} remain
  clearly worst at both budgets.
\item
  The seeded reruns confirm that the large penalties are not artifacts
  of a narrow anchor subset. At \texttt{5} and \texttt{10} minutes the
  full-screen evidence now shows \texttt{D}, \texttt{A}, \texttt{B}, and
  \texttt{C} retaining non-zero estimates after within-budget BH
  correction, while \texttt{E} is effectively null.
\item
  This changes the status of the \texttt{10}-minute story from
  "suggestive point estimates" to "corrected main-effect evidence with
  explicit seed uncertainty."
\end{itemize}

\subsection{5.9 60-Minute Bridge
Package}\label{59-60-minute-bridge-package}

We ran a \texttt{60}-minute bridge package over four predeclared anchor
conditions with four independent seeds each (\texttt{16} rows total):
the best screened \texttt{10}-minute extreme, the best seeded bridge
condition at \texttt{ASPECT\_RATIO=64}, the best \texttt{10}-minute
greedy incumbent, and a predeclared control outside the reduced
low-penalty region.

Role-to-condition mapping:

\begin{itemize}
\tightlist
\item
  \texttt{bridge\_best}: \texttt{best\_bridge\_10min\_d8\_ar64}
\item
  \texttt{greedy\_winner}: \texttt{greedy\_winner\_10min\_s15}
\item
  \texttt{screened\_best}: \texttt{best\_screened\_10min\_c07}
\item
  \texttt{control}: \texttt{predeclared\_control\_c10}
\end{itemize}

\begin{longtable}[]{@{}lrrl@{}}
\toprule
role & 10 min mean & 60 min mean & 95\% CI \\
\midrule
\endhead
\bottomrule
\endlastfoot
\texttt{bridge\_best} & \texttt{1.096292} & \texttt{0.974511} &
\texttt{{[}0.972148,\ 0.976874{]}} \\
\texttt{greedy\_winner} & \texttt{1.090276} & \texttt{0.981837} &
\texttt{{[}0.979531,\ 0.984143{]}} \\
\texttt{screened\_best} & \texttt{1.088859} & \texttt{0.984299} &
\texttt{{[}0.983023,\ 0.985574{]}} \\
\texttt{control} & \texttt{1.140270} & \texttt{1.001604} &
\texttt{{[}0.999867,\ 1.003341{]}} \\
\end{longtable}

Descriptive cross-seed win counts in the same package:

\begin{itemize}
\tightlist
\item
  \texttt{bridge\_best\ \textless{}\ greedy}: \texttt{16/16}
\item
  \texttt{screened\_best\ \textless{}\ greedy}: \texttt{1/16}
\item
  \texttt{bridge\_best\ \textless{}\ control}: \texttt{16/16}
\item
  \texttt{greedy\ \textless{}\ control}: \texttt{16/16}
\item
  \texttt{bridge\_best\ -\ greedy} mean gap: \texttt{-0.007326}
  \texttt{val\_bpb} (standard error of the difference \texttt{0.001038},
  about \texttt{7.1} standard errors)
\end{itemize}

Figure 3 shows the anchor-level seed trajectories for this package and
makes the longer-horizon crossover visually explicit: the centered-width
bridge becomes the best mean \texttt{60}-minute anchor while the
predeclared control remains worst.

\begin{figure}
\centering
\includegraphics[width=\linewidth,height=0.82\textheight,keepaspectratio]{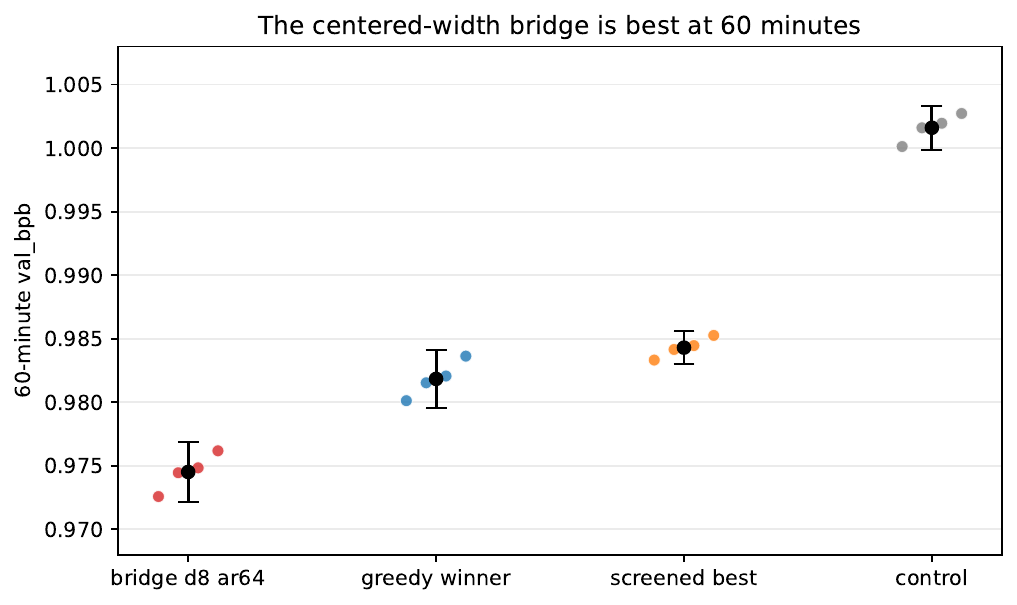}
\caption{The \texttt{60}-minute bridge package keeps the reduced-space
bridge best and the predeclared control worst. Points show seed runs;
black markers show means with 95\% confidence intervals.}
\end{figure}

Interpretation:

\begin{itemize}
\tightlist
\item
  The predeclared control remains worst at \texttt{60\ min}, so the
  reduced-space story does not collapse at the longer horizon.
\item
  The best centered-width bridge condition becomes the best mean
  \texttt{60}-minute performer and beats the greedy winner in all
  descriptive cross-seed win counts in this package.
\item
  The best screened extreme remains the best mean condition at
  \texttt{10} minutes in the full seeded-screen reruns, but it is
  overtaken by the centered-width bridge at \texttt{60} minutes. This is
  exactly the kind of transition the staged workflow is meant to expose.
\item
  The \texttt{60}-minute winner is also the larger bridge model
  (\texttt{depth=8}, about \texttt{50.3M} parameters), while the
  screened best extreme and greedy winner are both \texttt{depth=6}
  configurations (\texttt{39.8M} parameters). This is consistent with
  the paper\textquotesingle s budget-relaxation story, but this anchor
  package does not by itself separate a workflow effect from a
  later-horizon capacity effect inside the bridge region.
\item
  The \texttt{60}-minute package is therefore consistent with both local
  workflow refinement and simple capacity relaxation; it should not be
  read as isolating a workflow-only effect.
\item
  The best screened extreme remains clearly better than the control, but
  no longer matches the bridge or greedy winner. Within this anchor
  package, that pattern is consistent with keeping a local
  bridge-refinement step rather than freezing the original screened
  extreme as the final answer.
\item
  This is still an anchor-set continuation check, not a claim of global
  long-horizon optimality.
\end{itemize}

For scale, the \texttt{60}-minute bridge-vs-greedy mean gap is
\texttt{0.007326} \texttt{val\_bpb}, about \texttt{21\%} of the seeded
\texttt{10}-minute \texttt{D} main-effect penalty and about
\texttt{41\%} of the seeded \texttt{10}-minute \texttt{A} penalty. This
makes the bridge gap operationally visible inside the
paper\textquotesingle s metric, while still leaving downstream-transfer
value untested.

\subsection{5.10 Cross-Host Linux Anchor
Replication}\label{510-cross-host-linux-anchor-replication}

We then reran the same four-anchor package on a Linux host with one
NVIDIA L40S GPU at \texttt{10} and \texttt{60} minutes with four seeds
per anchor (\texttt{32} rows total). This is not a like-for-like A100
replication. It is a bounded cross-host check asking whether the
strongest operational signal from the original host remains visible
after a change in operating system, runtime path, and accelerator class.

Linux condition ordering:

For \texttt{10\ min}:

\begin{itemize}
\tightlist
\item
  best mean: \texttt{best\_bridge\_10min\_d8\_ar64} = \texttt{1.147247}
  {[}\texttt{1.144626}, \texttt{1.149868}{]}
\item
  second mean: \texttt{predeclared\_control\_c10} = \texttt{1.156941}
  {[}\texttt{1.155960}, \texttt{1.157922}{]}
\item
  worst mean: \texttt{best\_screened\_10min\_c07} = \texttt{1.175117}
  {[}\texttt{1.166961}, \texttt{1.183273}{]}
\end{itemize}

For \texttt{60\ min}:

\begin{itemize}
\tightlist
\item
  best mean: \texttt{best\_bridge\_10min\_d8\_ar64} = \texttt{1.003547}
  {[}\texttt{1.002316}, \texttt{1.004777}{]}
\item
  second mean: \texttt{predeclared\_control\_c10} = \texttt{1.011303}
  {[}\texttt{1.009919}, \texttt{1.012686}{]}
\item
  worst mean: \texttt{best\_screened\_10min\_c07} = \texttt{1.036137}
  {[}\texttt{1.033837}, \texttt{1.038437}{]}
\end{itemize}

Descriptive cross-seed win counts on Linux:

\begin{itemize}
\tightlist
\item
  \texttt{bridge\_best\ \textless{}\ greedy}: \texttt{16/16} at
  \texttt{10\ min}, \texttt{16/16} at \texttt{60\ min}
\item
  \texttt{bridge\_best\ \textless{}\ control}: \texttt{16/16} at
  \texttt{10\ min}, \texttt{16/16} at \texttt{60\ min}
\item
  \texttt{greedy\ \textless{}\ control}: \texttt{0/16} at
  \texttt{10\ min}, \texttt{0/16} at \texttt{60\ min}
\end{itemize}

Interpretation:

\begin{itemize}
\tightlist
\item
  The strongest part of the Windows story remains after the host change:
  the bridge has the lowest sample mean and continues to beat the greedy
  anchor at both budgets.
\item
  The full original anchor ordering does not survive unchanged. On
  Linux, the predeclared control does not remain worst; it ranks second
  at both budgets, while the screened extreme becomes worst.
\item
  The cross-host evidence is therefore mixed rather than fully
  confirmatory. It supports bridge-centered continuation evidence, but
  it does not support a stronger claim that the same full anchor
  hierarchy or bad-control separation is host-invariant.
\end{itemize}

\subsection{5.11 Twelve-Hour Three-Anchor Continuations on Windows and
Linux}\label{511-twelve-hour-three-anchor-continuations-on-windows-and-linux}

To test whether the strongest bridge-centered signal remains visible
beyond \texttt{60} minutes, we then ran the same three-anchor
continuation (\texttt{bridge}, \texttt{greedy}, \texttt{control}) at
\texttt{12h} on both hosts with two seeds per anchor (\texttt{12} rows
total across the two hosts). These later anchor packages are small
descriptive sample-mean checks, not powered inferential studies. Their
Student-\texttt{t} intervals use the within-cell replicate count
(\texttt{n=2}, hence \texttt{df=1}) and are shown only as coarse
dispersion summaries.

Windows \texttt{12h} condition ordering:

\begin{itemize}
\tightlist
\item
  best mean: \texttt{best\_bridge\_10min\_d8\_ar64} = \texttt{0.926420}
  {[}\texttt{0.914934}, \texttt{0.937906}{]}
\item
  second mean: \texttt{greedy\_winner\_10min\_s15} = \texttt{0.941248}
  {[}\texttt{0.935511}, \texttt{0.946984}{]}
\item
  worst mean: \texttt{predeclared\_control\_c10} = \texttt{0.954340}
  {[}\texttt{0.948615}, \texttt{0.960064}{]}
\end{itemize}

Linux \texttt{12h} condition ordering:

\begin{itemize}
\tightlist
\item
  best mean: \texttt{best\_bridge\_10min\_d8\_ar64} = \texttt{0.937047}
  {[}\texttt{0.934620}, \texttt{0.939474}{]}
\item
  second mean: \texttt{predeclared\_control\_c10} = \texttt{0.956076}
  {[}\texttt{0.953052}, \texttt{0.959100}{]}
\item
  worst mean: \texttt{greedy\_winner\_10min\_s15} = \texttt{0.961561}
  {[}\texttt{0.949668}, \texttt{0.973454}{]}
\end{itemize}

Descriptive cross-seed win counts at \texttt{12h}:

\begin{itemize}
\tightlist
\item
  Windows:

  \begin{itemize}
  \tightlist
  \item
    \texttt{bridge\_best\ \textless{}\ greedy}: \texttt{4/4}
  \item
    \texttt{bridge\_best\ \textless{}\ control}: \texttt{4/4}
  \item
    \texttt{greedy\ \textless{}\ control}: \texttt{4/4}
  \end{itemize}
\item
  Linux:

  \begin{itemize}
  \tightlist
  \item
    \texttt{bridge\_best\ \textless{}\ greedy}: \texttt{4/4}
  \item
    \texttt{bridge\_best\ \textless{}\ control}: \texttt{4/4}
  \item
    \texttt{greedy\ \textless{}\ control}: \texttt{0/4}
  \end{itemize}
\end{itemize}

Interpretation:

\begin{itemize}
\tightlist
\item
  The strongest operational signal remains visible at \texttt{12h} on
  both hosts: the bridge has the lowest sample mean and continues to
  beat the greedy anchor in descriptive pairwise counts.
\item
  Windows \texttt{12h} preserves the full
  \texttt{bridge\ \textless{}\ greedy\ \textless{}\ control} ordering
  seen in the main-host \texttt{60}-minute story, so the same-host
  long-horizon reading is materially stronger than it was at \texttt{60}
  minutes alone.
\item
  Linux \texttt{12h} again preserves the bridge-centered advantage but
  not the rest of the hierarchy. The control does not remain worst; it
  ranks above the greedy anchor.
\item
  The combined \texttt{12h} read therefore strengthens the paper only in
  a bounded descriptive sense: the bridge has the lowest sample mean in
  both small three-anchor packages, while the rest of the ordering
  remains host-sensitive.
\end{itemize}

\subsection{5.12 Twenty-Four-Hour Three-Anchor Continuations on Windows
and
Linux}\label{512-twenty-four-hour-three-anchor-continuations-on-windows-and-linux}

We then extended the same three-anchor continuation (\texttt{bridge},
\texttt{greedy}, \texttt{control}) to \texttt{24h} on both hosts with
two seeds per anchor (\texttt{12} rows total across the two hosts).
These full-day packages remain small and descriptive; they test whether
the bridge keeps the lowest sample mean at the longest horizon in this
paper, not whether the ranking is established with high-power
long-horizon inference.

Windows \texttt{24h} condition ordering:

\begin{itemize}
\tightlist
\item
  best mean: \texttt{best\_bridge\_10min\_d8\_ar64} = \texttt{0.923374}
  {[}\texttt{0.910319}, \texttt{0.936430}{]}
\item
  second mean: \texttt{greedy\_winner\_10min\_s15} = \texttt{0.938600}
  {[}\texttt{0.933867}, \texttt{0.943334}{]}
\item
  worst mean: \texttt{predeclared\_control\_c10} = \texttt{0.950929}
  {[}\texttt{0.943623}, \texttt{0.958235}{]}
\end{itemize}

Linux \texttt{24h} condition ordering:

\begin{itemize}
\tightlist
\item
  best mean: \texttt{best\_bridge\_10min\_d8\_ar64} = \texttt{0.930297}
  {[}\texttt{0.926377}, \texttt{0.934216}{]}
\item
  second mean: \texttt{predeclared\_control\_c10} = \texttt{0.952079}
  {[}\texttt{0.950758}, \texttt{0.953400}{]}
\item
  worst mean: \texttt{greedy\_winner\_10min\_s15} = \texttt{0.954650}
  {[}\texttt{0.941715}, \texttt{0.967585}{]}
\end{itemize}

Descriptive cross-seed win counts at \texttt{24h}:

\begin{itemize}
\tightlist
\item
  Windows:

  \begin{itemize}
  \tightlist
  \item
    \texttt{bridge\_best\ \textless{}\ greedy}: \texttt{4/4}
  \item
    \texttt{bridge\_best\ \textless{}\ control}: \texttt{4/4}
  \item
    \texttt{greedy\ \textless{}\ control}: \texttt{4/4}
  \end{itemize}
\item
  Linux:

  \begin{itemize}
  \tightlist
  \item
    \texttt{bridge\_best\ \textless{}\ greedy}: \texttt{4/4}
  \item
    \texttt{bridge\_best\ \textless{}\ control}: \texttt{4/4}
  \item
    \texttt{greedy\ \textless{}\ control}: \texttt{0/4}
  \end{itemize}
\end{itemize}

Interpretation:

\begin{itemize}
\tightlist
\item
  The bridge-refined anchor has the lowest sample mean at \texttt{24h}
  on both hosts and beats both the greedy anchor and the predeclared
  control in all descriptive cross-seed pairings.
\item
  Windows \texttt{24h} preserves the same
  \texttt{bridge\ \textless{}\ greedy\ \textless{}\ control} ordering
  already seen at \texttt{12h}, so the same-host long-horizon read
  remains internally consistent through a full-day horizon.
\item
  Linux \texttt{24h} again preserves the bridge-centered advantage but
  not the rest of the hierarchy. The control does not remain worst; it
  ranks above the greedy anchor.
\item
  The \texttt{24h} read therefore strengthens the paper in one bounded
  descriptive way: the bridge has the lowest sample mean in both small
  three-anchor packages through \texttt{24h}, while the non-bridge
  ordering remains host-sensitive.
\end{itemize}

Figure 4 combines the four long-horizon anchor packages and makes the
bounded cross-host pattern visible at a glance: Windows preserves
\texttt{bridge\ \textless{}\ greedy\ \textless{}\ control} at both
horizons, while Linux preserves the bridge as best but flips the
non-bridge ordering.

\begin{figure}
\centering
\includegraphics[width=\linewidth,height=0.82\textheight,keepaspectratio]{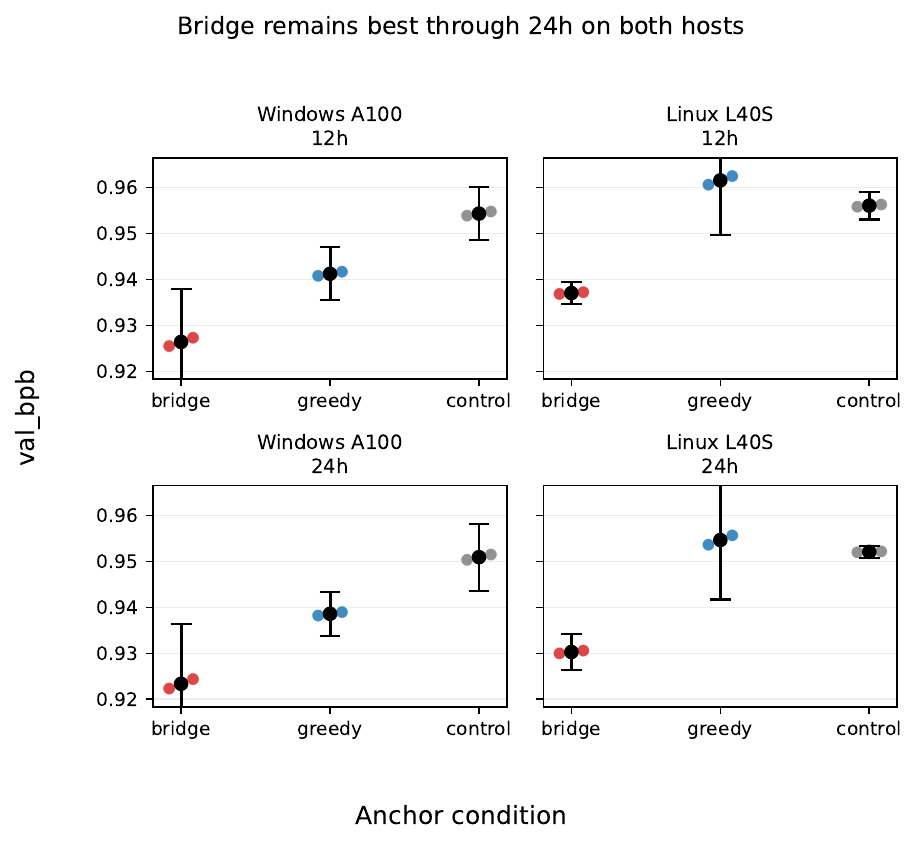}
\caption{Dual-host long-horizon anchor packages at \texttt{12h} and
\texttt{24h}. In all four panels, the bridge anchor has the lowest
sample mean. Windows preserves
\texttt{bridge\ \textless{}\ greedy\ \textless{}\ control} at both
horizons, while Linux ranks \texttt{control\ \textless{}\ greedy} among
the two non-bridge anchors. Points show seed runs; black markers show
means with 95\% confidence intervals.}
\end{figure}

\section{6. Discussion}\label{6-discussion}

The evidence supports a bounded methodological conclusion. In this host
regime, short-budget training behavior is dominated by a small set of
main penalties, especially total batch and model size. Those penalties
relax substantially as budget increases from \texttt{2} to \texttt{10}
minutes. The new full seeded-screen reruns sharpen that claim
materially. At \texttt{5} and \texttt{10} minutes, the seeded
fixed-effects models show \texttt{D}, \texttt{A}, \texttt{B}, and
\texttt{C} retaining non-zero estimates once the full
\texttt{16}-condition design is rerun across five independent seeds per
condition and the within-budget BH correction is applied, while
\texttt{E} does not. That is the cleanest answer to the earlier concern
that the \texttt{10}-minute effects might be at the noise floor: some of
them are, but not all of them, and the full seeded reruns now separate
the real effects from the null one.

The targeted seeded anchor package remains useful because it covers the
\texttt{2}-minute regime and the centered-width bridge conditions
directly. In that subset, budget and condition structure dominate seed
identity, worst regions remain clearly separated from best and bridge
regions, and the bridge ordering shifts with budget exactly where the
operational refinement story says it should. The later
\texttt{60}-minute bridge package then sharpens the longer-horizon
interpretation. The best screened extreme does not remain best at the
longer horizon; the centered-width bridge does. That is important. The
recommendation is therefore not "screen once and stop." The
recommendation is "screen, confirm, then refine locally." The
matched-cost random baseline sharpens the same point from the other
side. In this \texttt{32}-condition space, random search can hit a
strong incumbent. What it does not do is explain why that incumbent is
good or why the good incumbents cluster in the same low-penalty region.
The practical value of the design is interpretability and search-space
reduction, not a broad claim that designed screening universally
outperforms greedy or random search. A greedy trajectory can still
locate a strong incumbent, and a random batch can sometimes do so as
well. What the present study shows is narrower: best-so-far trajectories
alone do not provide direct factor attribution, whereas a compact staged
screen can identify high-penalty directions and expose a plausible
refinement region before larger search budgets are spent. The
\texttt{60}-minute bridge package should also be read narrowly: it shows
that the staged workflow surfaces a reduced region in which a larger
centered-width bridge becomes the best anchor, but it does not fully
disentangle workflow value from the later-horizon capacity advantage of
that bridge model.

The new Linux anchor package changes the paper\textquotesingle s
external-validity story, but only partially. At \texttt{10} minutes,
\texttt{60} minutes, \texttt{12h}, and now \texttt{24h}, the strongest
anchor-level result remains visible cleanly: on Linux, the bridge has
the lowest sample mean and continues to beat the greedy anchor. That
weakens the earlier single-host caveat in a meaningful way because the
bridge-centered result is no longer confined to the original Windows
A100 runtime path or to the shorter continuation horizons. At the same
time, the Linux packages do not preserve the full original anchor
ordering. The predeclared control does not remain worst at \texttt{10}
minutes, \texttt{60} minutes, \texttt{12h}, or \texttt{24h}, and the
rest of the ranking remains host-sensitive. This means the paper should
not claim full cross-host replication or host-invariant preservation of
the complete ranking structure. The correct reading is narrower: the
bridge-centered sample-mean result is directionally stable across hosts
through \texttt{24h}, while other anchor relationships are
host-sensitive.

This workflow should therefore be read as a bounded method for
short-horizon experimentation with limited longer-horizon support on a
small anchor set and cross-host support for the bridge-centered
refinement step through \texttt{24h}. It does not establish global
long-horizon optimality, hardware-invariant ranking structure, or
scaling-law generality. Those require additional experiments beyond the
present bounded regime.

\section{7. Limitations}\label{7-limitations}

\begin{enumerate}
\def\labelenumi{\arabic{enumi}.}
\tightlist
\item
  Cross-host evidence is still partial rather than complete. The paper
  now includes bounded Linux L40S anchor packages at \texttt{10}
  minutes, \texttt{60} minutes, \texttt{12h}, and \texttt{24h}, but the
  original full anchor ordering does not survive unchanged across hosts
  and the results do not support a hardware-invariant claim.
\item
  Focus on final \texttt{val\_bpb}, not downstream task transfer
  metrics. No downstream evaluation is included, so practitioner
  recommendations are provisional for validation-compression performance
  in this runner rather than demonstrated task-transfer guidance.
\item
  The \texttt{60}-minute, \texttt{12h}, and \texttt{24h} packages are
  still anchor-set continuation checks over a small predeclared anchor
  set; the \texttt{12h} and \texttt{24h} packages have only \texttt{n=2}
  seeds per anchor and should be read as coarse descriptive sample-mean
  hardening, not high-power long-horizon inference.
\item
  Search baselines are still bounded to same-host greedy and a
  single-budget random-search stress test; the paper does not include
  adaptive multi-fidelity baselines such as Hyperband, BOHB, ASHA, or
  population-based training.
\item
  The branch now includes full \texttt{16}-condition seeded screens at
  \texttt{5} and \texttt{10} minutes plus a targeted seeded anchor
  subset across all three budgets, but the full \texttt{2}-minute
  screen, greedy baselines, and the complete D-fixed grid were not rerun
  across broad seed distributions.
\item
  Reproducibility is now package-explicit and archive-frozen at the
  snapshot level, but still not ideal from a software-distribution
  perspective: the study depends on a frozen local source snapshot plus
  matrix artifacts rather than a single tagged public repository for the
  entire paper workspace.
\item
  The work is methodological and small-scale, with no human-subject data
  or deployed system evaluation, but it still reduces the cost of
  iterative recipe search for language-model training. As with other
  training-efficiency methods, that can have dual-use value if
  repurposed for broader capability-seeking automation. We therefore
  present the contribution as an auditable bounded workflow result
  rather than a turnkey optimization system.
\end{enumerate}

\section{8. Conclusion}\label{8-conclusion}

This paper started from an interaction-atlas hypothesis and ends with a
narrower claim: under strict short budgets on the main host,
training-recipe behavior is dominated by budget-sensitive penalties.
Three-budget screens identify the large penalties early, full
seeded-screen reruns at \texttt{5} and \texttt{10} minutes show that
\texttt{D}, \texttt{A}, \texttt{B}, and \texttt{C} retain non-zero
estimates after within-budget correction while \texttt{E} does not, the
matched-cost random baseline shows that strong incumbents can be reached
by chance in this small space but mostly inside the same low-penalty
region, rerun-complete D-fixed follow-ups show that interactions remain
real but secondary in absolute size inside the legacy fixed-seed regime,
the targeted multi-seed subset shows that worst regions remain worse
across seeds while centered-width bridges become better refinement
targets at \texttt{5} and \texttt{10} minutes, the later
\texttt{60}-minute bridge package shows that the best centered-width
bridge remains the strongest anchor condition at a longer horizon, and
the later \texttt{12h} and \texttt{24h} continuations show that the
bridge has the lowest sample mean in these small descriptive
continuations on both hosts even though the full anchor ordering does
not.

The operational recommendation is therefore staged: screen to eliminate
high-penalty directions, confirm the anchor regions, then refine locally
inside the reduced space before spending larger automated search
budgets. The \texttt{60}-minute bridge result matters precisely because
it shows that the local refinement step can overtake both the original
screened extreme and the bounded greedy incumbent once the budget is
long enough for the depth penalty to relax. The later dual-host
continuation packages matter because they show that the bridge keeps the
lowest sample mean in these small descriptive three-anchor continuations
on both hosts through \texttt{24h}. The Linux replication matters
because it shows that this bridge-centered recommendation is not
confined to a single runtime path, even though the rest of the anchor
hierarchy still appears host-sensitive.

\section{Appendix A. Reproducibility
Snapshot}\label{appendix-a-reproducibility-snapshot}

The upstream \texttt{autoresearch} baseline used for this study was
frozen from a local clone at commit:

\begin{itemize}
\tightlist
\item
  \texttt{228791fb499afffb54b46200aca536f79142f117}
\end{itemize}

The runner-relevant local source snapshot used to materialize the
experiment engines in this paper workspace is archived as:

\begin{itemize}
\tightlist
\item
  \texttt{autoresearch-seed.zip}
\item
  SHA256:
  \texttt{D15B7F68F9BDFF6E06D58FB9E2692152D8B44C34CC34652435C1F94675ADCADC}
\item
  public contents manifest:
  \texttt{autoresearch\_seed\_snapshot\_manifest\_2026-04-27.json}
\end{itemize}

The local control and parsing artifacts that define the execution path
for this branch are:

\begin{itemize}
\tightlist
\item
  \texttt{materialize\_condition.py}
\item
  \texttt{parse\_train\_summary.py}
\item
  \texttt{patch\_prepare\_time\_budget.py}
\item
  \texttt{train\_windows\_fallback.py}
\item
  \texttt{remote-env.ps1}
\item
  \texttt{remote-env-linux.ps1}
\end{itemize}

The package-level configuration dumps for the major result blocks are
stored as explicit matrices and base-condition tables in the paper
workspace. The principal ones are:

\begin{itemize}
\tightlist
\item
  \texttt{pilot\_16run\_design\_matrix.csv}
\item
  \texttt{replication\_12run\_matrix.csv}
\item
  \texttt{followup\_abce\_dlow\_16run\_matrix.csv}
\item
  \texttt{followup\_abce\_dlow\_16run\_matrix\_10min.csv}
\item
  \texttt{center\_points\_bridge\_9run\_matrix.csv}
\item
  \texttt{seed\_confirmation\_base\_conditions\_6row.csv}
\item
  \texttt{full\_screen\_seeded\_base\_conditions\_16row.csv}
\item
  \texttt{bridge\_60min\_base\_conditions\_4row.csv}
\item
  \texttt{crosshost\_linux\_anchor\_base\_conditions\_4row.csv}
\item
  \texttt{long\_horizon\_anchor\_base\_conditions\_3row.csv}
\item
  \texttt{long\_horizon\_anchor\_12h\_6run.csv}
\item
  \texttt{long\_horizon\_anchor\_24h\_base\_conditions\_3row.csv}
\item
  \texttt{long\_horizon\_anchor\_24h\_6run.csv}
\item
  \texttt{table1\_run\_manifest\_2026-04-27.csv}
\item
  \texttt{table1\_run\_manifest\_2026-04-27.json}
\end{itemize}

The CSV manifest contains one package row per Table 1 block. Aggregate
totals live in the companion JSON manifest so that the CSV cannot be
naively double-counted.

The fixed data/tokenizer identifiers used by the runner are:

\begin{itemize}
\tightlist
\item
  dataset base URL:
  \url{https://huggingface.co/datasets/karpathy/climbmix-400b-shuffle/resolve/main}
\item
  pinned validation shard: \texttt{shard\_06542.parquet}
\item
  tokenizer artifacts expected by the runner: \texttt{tokenizer.pkl} and
  \texttt{token\_bytes.pt}
\item
  tokenizer vocabulary size: \texttt{8192}
\end{itemize}

Together, the source snapshot, generator code, dataset and tokenizer
identifiers, package matrices, and curated CSV/JSON summaries provide
the configuration-level audit trail for every package reported in Table
1.

\section{References}\label{references}

{[}1{]} Andrej Karpathy. \emph{autoresearch}. GitHub repository.
Available at: \url{https://github.com/karpathy/autoresearch}

{[}2{]} SkyPilot documentation. \emph{Parallel autoresearch}. Available
at:
\url{https://docs.skypilot.co/en/latest/examples/agents/autoresearch.html}

{[}3{]} \emph{Predictable Scale: Part I, Step Law -\/- Optimal
Hyperparameter Scaling Law in Large Language Model Pretraining}.
arXiv:2503.04715. \url{https://arxiv.org/abs/2503.04715}

{[}4{]} \emph{Principled Architecture-aware Scaling of Hyperparameters}.
arXiv:2402.17440. \url{https://arxiv.org/abs/2402.17440}

{[}5{]} \emph{Rethinking Language Model Scaling under Transferable
Hypersphere Optimization}.arXiv:2603.28743.
\url{https://arxiv.org/abs/2603.28743}

{[}6{]} \emph{Practical Efficiency of Muon for Pretraining}.
arXiv:2505.02222. \url{https://arxiv.org/abs/2505.02222}

{[}7{]} \emph{Hyperparameter Transfer Enables Consistent Gains of
Matrix-Preconditioned Optimizers Across Scales}. arXiv:2512.05620.
\url{https://arxiv.org/abs/2512.05620}

{[}8{]} \emph{AdaMuon: Adaptive Muon Optimizer}. arXiv:2507.11005.
\url{https://arxiv.org/abs/2507.11005}

{[}9{]} James Bergstra and Yoshua Bengio. \emph{Random Search for
Hyper-Parameter Optimization}. Journal of Machine Learning Research,
13(10):281-305, 2012.
\url{https://jmlr.org/beta/papers/v13/bergstra12a.html}

{[}10{]} Frank Hutter, Holger H. Hoos, and Kevin Leyton-Brown. \emph{An
Efficient Approach for Assessing Hyperparameter Importance}. Proceedings
of the 31st International Conference on Machine Learning, PMLR 32, 2014.
\url{https://proceedings.mlr.press/v32/hutter14.html}

{[}11{]} Lisha Li, Kevin Jamieson, Giulia DeSalvo, Afshin Rostamizadeh,
and Ameet Talwalkar. \emph{Hyperband: A Novel Bandit-Based Approach to
Hyperparameter Optimization}. Journal of Machine Learning Research,
18(185):1-52, 2018. \url{https://jmlr.org/beta/papers/v18/16-558.html}

{[}12{]} Stefan Falkner, Aaron Klein, and Frank Hutter. \emph{BOHB:
Robust and Efficient Hyperparameter Optimization at Scale}. Proceedings
of the 35th International Conference on Machine Learning, PMLR 80, 2018.
\url{https://proceedings.mlr.press/v80/falkner18a.html}

{[}13{]} Jasper Snoek, Hugo Larochelle, and Ryan P. Adams.
\emph{Practical Bayesian Optimization of Machine Learning Algorithms}.
Advances in Neural Information Processing Systems 25, 2012.
\url{https://proceedings.neurips.cc/paper/2012/hash/05311655a15b75fab86956663e1819cd-Abstract.html}

{[}14{]} Douglas C. Montgomery. \emph{Design and Analysis of
Experiments}. Wiley, 10th edition, 2019.

{[}15{]} George E. P. Box, J. Stuart Hunter, and William G. Hunter.
\emph{Statistics for Experimenters: Design, Innovation, and Discovery}.
Wiley, 2nd edition, 2005.

{[}16{]} C. F. Jeff Wu and Michael Hamada. \emph{Experiments: Planning,
Analysis, and Optimization}. Wiley, 2nd edition, 2009.

{[}17{]} George E. P. Box and K. B. Wilson. \emph{On the Experimental
Attainment of Optimum Conditions}. Journal of the Royal Statistical
Society, Series B, 13(1):1-45, 1951.

{[}18{]} Raymond H. Myers, Douglas C. Montgomery, and Christine M.
Anderson-Cook. \emph{Response Surface Methodology: Process and Product
Optimization Using Designed Experiments}. Wiley, 4th edition, 2016.

{[}19{]} Jordan Hoffmann, Sebastian Borgeaud, Arthur Mensch, Elena
Buchatskaya, Trevor Cai, Eliza Rutherford, Diego de Las Casas, Lisa Anne
Hendricks, Johannes Welbl, Aidan Clark, Tom Hennigan, Eric Noland, Katie
Millican, George van den Driessche, Bogdan Damoc, Aurelia Guy, Simon
Osindero, Karen Simonyan, Erich Elsen, Jack W. Rae, Oriol Vinyals, and
Laurent Sifre. \emph{Training Compute-Optimal Large Language Models}.
arXiv:2203.15556, 2022. \url{https://arxiv.org/abs/2203.15556}

{[}20{]} Max Jaderberg, Valentin Dalibard, Simon Osindero, Wojciech M.
Czarnecki, Jeff Donahue, Ali Razavi, Oriol Vinyals, Tim Green, Iain
Dunning, Karen Simonyan, Chrisantha Fernando, and Koray Kavukcuoglu.
\emph{Population Based Training of Neural Networks}. arXiv:1711.09846,
2017. \url{https://arxiv.org/abs/1711.09846}

{[}21{]} Liam Li, Kevin Jamieson, Afshin Rostamizadeh, Ekaterina Gonina,
Moritz Hardt, Benjamin Recht, and Ameet Talwalkar. \emph{A System for
Massively Parallel Hyperparameter Tuning}. arXiv:1810.05934, 2018.
\url{https://arxiv.org/abs/1810.05934}

{[}22{]} Leslie N. Smith. \emph{Cyclical Learning Rates for Training
Neural Networks}. IEEE Winter Conference on Applications of Computer
Vision, 2017. arXiv:1506.01186. \url{https://arxiv.org/abs/1506.01186}

{[}23{]} Leslie N. Smith and Nicholay Topin. \emph{Super-Convergence:
Very Fast Training of Neural Networks Using Large Learning Rates}.
Proceedings of SPIE 11006, Artificial Intelligence and Machine Learning
for Multi-Domain Operations Applications, 2019. arXiv:1708.07120.
\url{https://arxiv.org/abs/1708.07120}

{[}24{]} Ashish Vaswani, Noam Shazeer, Niki Parmar, Jakob Uszkoreit,
Llion Jones, Aidan N. Gomez, Lukasz Kaiser, and Illia Polosukhin.
\emph{Attention Is All You Need}. arXiv:1706.03762, 2017.
\url{https://arxiv.org/abs/1706.03762}

\end{document}